\def\year{2022}\relax
\documentclass[letterpaper]{article} 
\usepackage{aaai22}  
\usepackage{times}  
\usepackage{helvet}  
\usepackage{courier}  
\usepackage[hyphens]{url}  
\usepackage{graphicx} 
\usepackage{natbib}  
\usepackage{caption} 
\DeclareCaptionStyle{ruled}{labelfont=normalfont,labelsep=colon,strut=off} 
\usepackage{algorithm}
\usepackage{algorithmic}
\usepackage{booktabs}
\usepackage{rotating}
\usepackage{makecell}
\usepackage{amssymb}
\usepackage{subfigure}
\usepackage{url}
\usepackage[colorlinks, linkcolor=blue]{hyperref}

\usepackage{newfloat}
\usepackage{listings}
\floatstyle{ruled}
\newfloat{listing}{tb}{lst}{}
\floatname{listing}{Listing}
\title{WUDA: Unsupervised Domain Adaptation Based on Weak Source Domain Labels}
\author{
    Shengjie Liu\textsuperscript{\rm 1},
    Chuang Zhu\thanks{Corresponding author.}\textsuperscript{\rm 1},
    Wenqi Tang\textsuperscript{\rm 1}\\
}
\affiliations{
    \textsuperscript{\rm 1}Beijing University of Posts and Telecommunications, Beijing, China\\

    \{shengjie.Liu, czhu, tangwenqi\}@bupt.edu.cn
}

\usepackage{bibentry}

\begin{document}

\maketitle

\begin{abstract}
Unsupervised domain adaptation (UDA) for semantic segmentation addresses the cross-domain problem with fine source domain labels. However, the acquisition of semantic labels has always been a difficult step, many scenarios only have weak labels (e.g. bounding boxes). For scenarios where weak supervision and cross-domain problems coexist, this paper defines a new task: unsupervised domain adaptation based on weak source domain labels (WUDA). To explore solutions for this task, this paper proposes two intuitive frameworks: 1) Perform weakly supervised semantic segmentation in the source domain, and then implement unsupervised domain adaptation; 2) Train an object detection model using source domain data, then detect objects in the target domain and implement weakly supervised semantic segmentation. We observe that the two frameworks behave differently when the datasets change. Therefore, we construct dataset pairs with a wide range of domain shifts and conduct extended experiments to analyze the impact of different domain shifts on the two frameworks. In addition, to measure domain shift, we apply the metric representation shift to urban landscape image segmentation for the first time. The source code and constructed datasets are available at \url{https://github.com/bupt-ai-cz/WUDA}.
\end{abstract}

\section{Introduction}

As one of the most popular computer vision technologies, semantic segmentation has developed very mature and is widely used in many scenarios such as autonomous driving, remote sensing image recognition, and medical image processing. However, like most deep learning models, a serious problem faced by semantic segmentation models in the application is the existence of domain shift. Since the data distribution of the target domain may be very different from the source domain, a model that performs well in the source domain may experience catastrophic performance degradation in the target domain. For cross-domain problems, many studies on domain adaptation have improved the generalization ability of the model. Among these studies, unsupervised domain adaptation (UDA) for semantic segmentation can fine-tune a trained model by using self-training, adversarial training, or image style transfer without any additional annotations of the target domain. Methods of self-training \cite{zou2018unsupervised,Zou_2019_ICCV} and adversarial training \cite{tsai2018learning,luo2019taking} can adapt the model to the distribution of the target domain images. Image style transfer methods \cite{yang2020fda,yang2020label} can bring the target domain data closer to the distribution of the source domain. 

In deep learning tasks, massive amounts of samples are required for training to improve the robustness of the model, therefore, the annotation of large-scale datasets is another difficulty in training deep models. In the semantic segmentation task, in order to obtain pixel-wise mask annotations, it takes a lot of time for a single image (e.g. it takes 1.5 hours to label one image in the Cityscapes \cite{cordts2016cityscapes} dataset), and the annotation of the entire dataset requires huge manpower. Therefore, many computer vision datasets have only weak annotations (e.g. bounding boxes, points etc.)
\begin{figure}[t]
    \centering
    \includegraphics[width=0.99\columnwidth]{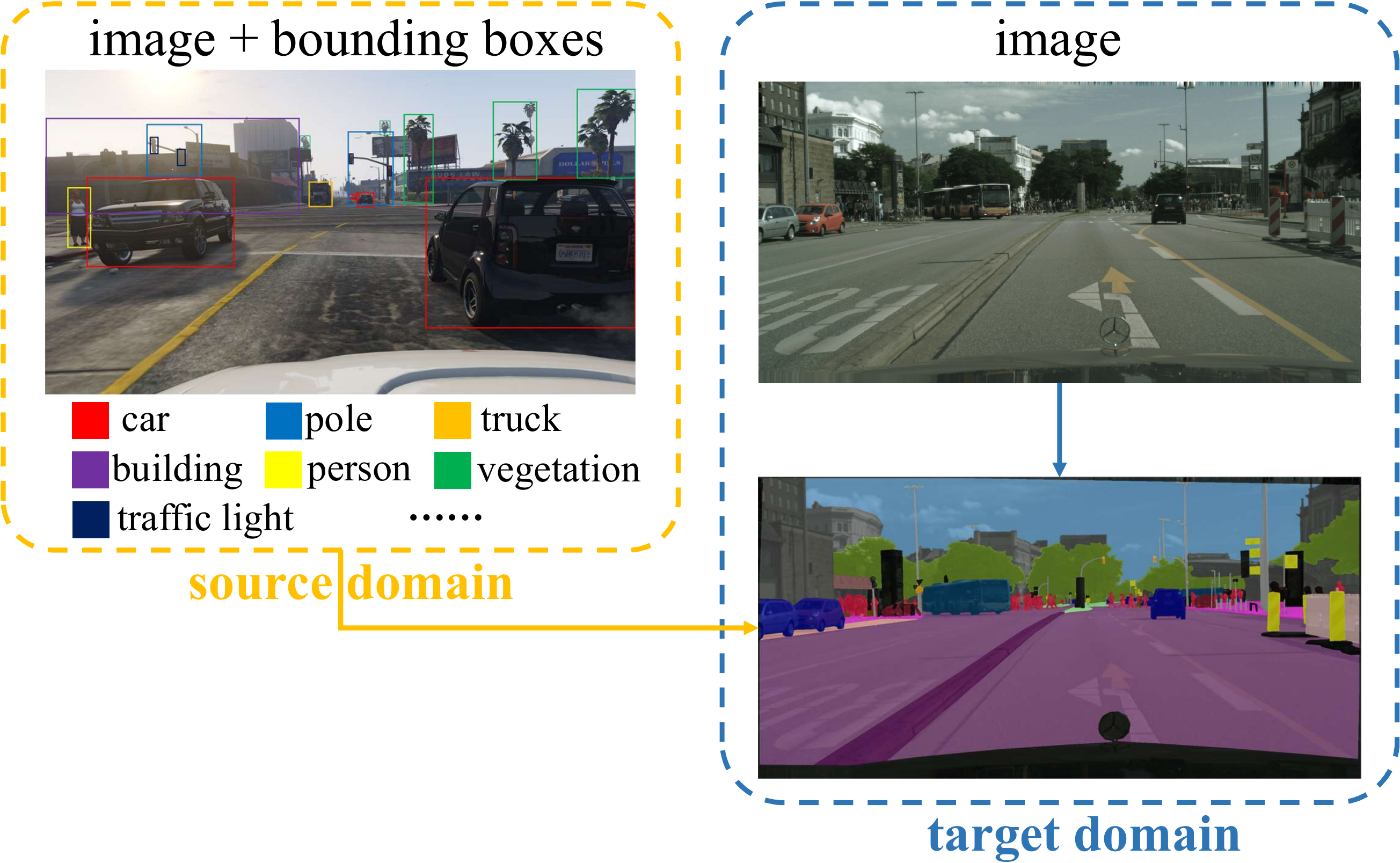} 
    \caption{Schematic diagram of WUDA. The source domain has only bounding boxes and the target domain has no annotations. WUDA achieves semantic segmentation of the target domain images under these conditions.}
    \label{task_show}
    \vspace{-1.0em}
\end{figure}

When the cross-domain problem and the weak labels coexist (only source domain bounding boxes and target domain images are available), the domain shift and the weak supervision both bring a negative contribution to the pixel-level semantic segmentation. In this case, it is a challenge to achieve accurate semantic segmentation in the target domain. We define this task as Unsupervised Domain Adaptation Based on Weak Source Domain Labels (WUDA). The schematic diagram of WUDA is shown in Figure~\ref{task_show}. For the newly defined task, it is necessary to explore a framework to tackle the problems of transfer learning and weakly supervised segmentation at the same time. The realization of this task can reduce the requirements for source domain labels in future UDA tasks.

In summary, this paper makes the following contributions:

\begin{itemize} 
    \item We define a novel task: unsupervised domain adaptation based on weak source domain labels (WUDA). For this task, we propose two intuitive frameworks: Weakly Supervised Semantic Segmentation + Unsupervised Domain Adaptation (WSSS-UDA) and Target Domain Object Detection + Weakly Supervised Semantic Segmentation (TDOD-WSSS).
    
    \item We benchmark typical weakly supervised semantic segmentation, unsupervised domain adaptation, and object detection techniques under our two proposed frameworks, and find that the results of framework WSSS-UDA can reach 83\% of the UDA method with fine source domain labels. 
    
    \item We construct a series of datasets with different domain shifts. To the best of our knowledge, we are the first to use representation shift for domain shift measurement in urban landscape datasets. The constructed dataset will be open for research on WUDA/UDA under multiple domain shifts.
    
    \item To further analyze the impact of different degrees of domain shift on our proposed frameworks, we conduct extended experiments using our constructed datasets and find that framework TDOD-WSSS is more sensitive to changes in domain shift than framework WSSS-UDA.
\end{itemize}

\section{Related Work}

WUDA  will involve weakly supervised semantic segmentation, unsupervised domain adaptation, object detection, and the measure of domain shift techniques. In this section, we will review these related previous works.

\subsection{Weakly Supervised Semantic Segmentation}

In computer vision tasks, pixel-wise mask annotations takes far more time compared to weak annotations \cite{lin2014microsoft}, and the need for time-saving motivates weakly supervised semantic segmentation. Labels for weakly supervised segmentation can be bounding boxes, points, scribbles and image-level tags. 
Methods \cite{dai2015boxsup,khoreva2017simple,li2018weakly,song2019box,kulharia2020box2seg} using bounding boxes as supervision usually employ GrabCut \cite{rother2004interactive} or segment proposals techniques to get more accurate semantic labels and can achieve results close (95\% or even higher) to fully supervised methods. Point-supervisied and scribble-supervised methods \cite{bearman2016s,qian2019weakly,lin2016scribblesup,vernaza2017learning,tang2018normalized,tang2018regularized} 
take advantage of location and category information in annotations and achieve excellent segmentation results. Tag-supervised methods \cite{jiang2019integral,wang2020self,lee2021railroad,li2021pseudo} often use class activation mapping (CAM) \cite{zhou2016learning} algorithm to obtain localization maps of the main objects in the images.

\begin{figure*}[t]
\centering
\subfigure[Linear regression results for datasets with 16 categories. Pearson correlation: -0.874]{
\label{Fig.sub.1}
\includegraphics[width=0.32\textwidth]{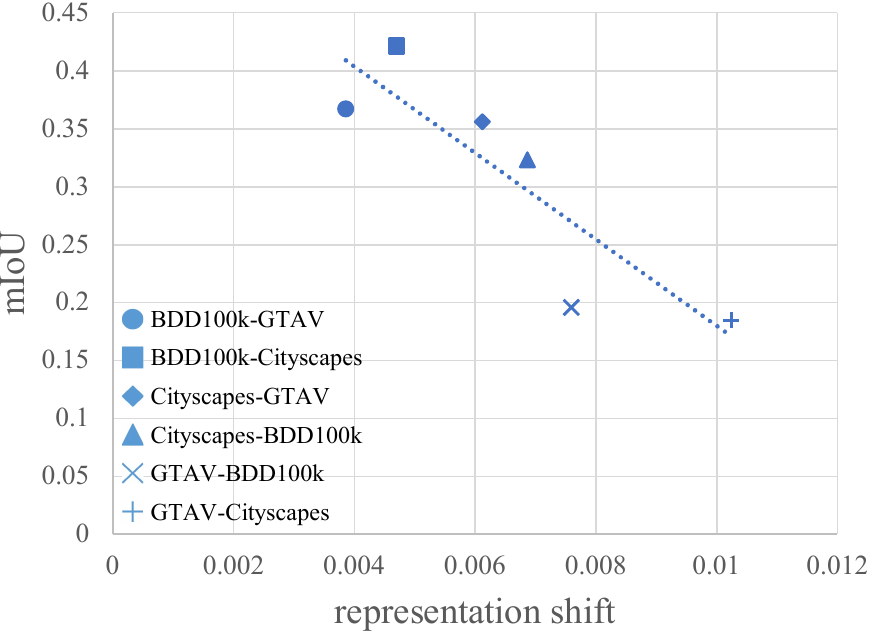}}
\subfigure[Linear regression results for datasets with 19 categories. Pearson correlation: -0.741]{
\label{Fig.sub.2}
\includegraphics[width=0.32\textwidth]{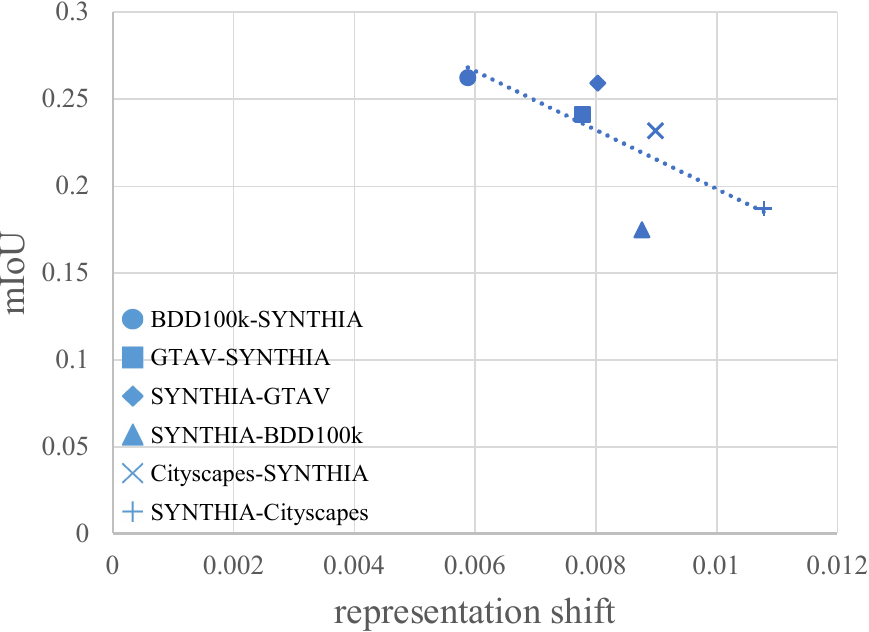}}
\subfigure[Linear regression results for augmented Cityscapes, the source domain dataset is GTAV. Pearson correlation: -0.936]{
\label{Fig.sub.3}
\includegraphics[width=0.32\textwidth]{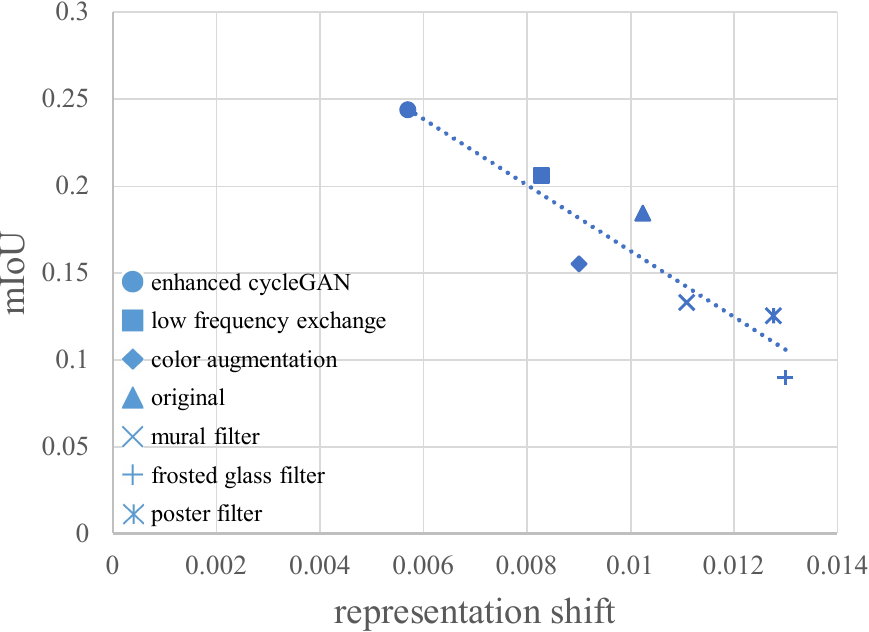}}
\caption{Linear regression results of target domain mIoU and representation shift. There is a strong correlation between the two metrics, demonstrating the feasibility of representation shift for semantic segmentation tasks.}
\label{analysis}
\vspace{-1.0em}
\end{figure*}
\subsection{Unsupervised Domain Adaptation for Semantic Segmentation}

Unsupervised Domain Adaptation (UDA) is committed to solving the problem of poor model generalization caused by inconsistent data distribution in the source and target domains. Self-training (ST) and adversarial training (AT) are key schemes of UDA: self-training schemes \cite{zou2018unsupervised,Zou_2019_ICCV,lian2019constructing,li2020content,lv2020cross,melas2021pixmatch,tranheden2021dacs,guo2021metacorrection,araslanov2021self} typically set a threshold to filter pseudo-labels with high confidence on the target domain, and use the pseudo-labels to supervise target domain training; adversarial training methods \cite{tsai2018learning,luo2019taking,luo2019significance,du2019ssf,vu2019advent,tsai2019domain,yang2020adversarial,wang2020classes,li2021bi} usually add a domain discriminator to the model. The adversarial game of the segmenter and the discriminator can make the segmentation results of the source and target domains tend to be consistent. There are also works \cite{zhang2019category,pan2020unsupervised,wang2020differential,yu2021dast,wang2021uncertainty,mei2020instance} that perform both self-training and adversarial training to achieve good segmentation results on the target domain.

\subsection{Object Detection}

Autonomous driving technology has greatly promoted the development of object detection. There are many pioneering works that can be widely used in various object detection tasks, such as some two-stage methods \cite{girshick14CVPR,girshickICCV15fastrcnn,renNIPS15fasterrcnn} that first perform object extraction, and then classify the extracted objects. Yolo series of algorithms \cite{redmon2016you,redmon2017yolo9000,redmon2018yolov3,bochkovskiy2020yolov4,glenn_jocher_2020_3983579} can simultaneously achieve object extraction and classification in one network. The current popular object detection method Yolov5 \cite{glenn_jocher_2020_3983579} has been able to achieve 72.7\% mean average precision (mAP) on the coco2017 val dataset. Object detection techniques also help to extract bounding boxes in weakly supervised segmentation methods \cite{lan2021discobox,lee2021bbam}.

\subsection{Domain Shift Assessment}

Domain shift comes from the difference between the source and target domain data. There are various factors (e.g. image content, view angle, image texture, etc.) that contribute to domain shift. While for Convolutional Neural Networks (CNN), texture is the most critical factor. Many studies \cite{geirhos2018imagenet,nam2019reducing} suggest that the focus of Convolutional Neural Networks (CNN) and human eyes is different when processing images: human eyes are sensitive to the content information of the image (e.g. shapes), while CNN is more sensitive to the style of the image (e.g. texture). Actually, if it involves the calculation of image texture differences, most methods are based on the output features of the middle layer of the neural network. For example, in work \cite{stacke2020measuring}, the metric representation shift is proposed to measure the domain shift of pathological images. In that paper, the mean value of each channel in the feature map is used as the style information of an image. The experimental results show that in the task of pathological image classification, the value of representation shift and the classification accuracy on the test set have a strong correlation, which justifies the use of the intermediate layer features to calculate the domain shift. In the pioneering work on image style transfer \cite{gatys2015texture}, the style loss is calculated using the Gram matrix of the output features of the two images.

\section{Construct Datasets with Different Domain Shifts}

In section \textbf{Experiments}, we will analyze the impact of domain shift changes on the experimental results. Therefore, we first introduce the domain shift metric $representation$ $shift$ for semantic segmentation and construct datasets with different representation shifts.

\subsection{Representation Shift for Semantic Segmentation}
Karin Stacke et al. \cite{stacke2020measuring} propose the metric representation shift to measure the domain shift between pathological image datasets. Since the capture of pathological images is affected by many factors (e.g. slide preparation, staining protocol, scanner properties, etc.), domain shift often exists between pathological images scanned at different medical centers. Therefore, for a pathological image classification model trained on the source domain, it may not achieve good results on test images. In work \cite{stacke2020measuring}, representation shift is highly correlated with the classification accuracy in the target domain: a high representation shift means that the accuracy on the target domain tends to be low. The metric solves this problem: in the absence of ground truth in the target domain, we can estimate whether the predictions made by the trained model are credible by computing the representation shift between the source and target domains. 

\begin{figure*}[t]
    \centering
    \includegraphics[width=0.99\textwidth]{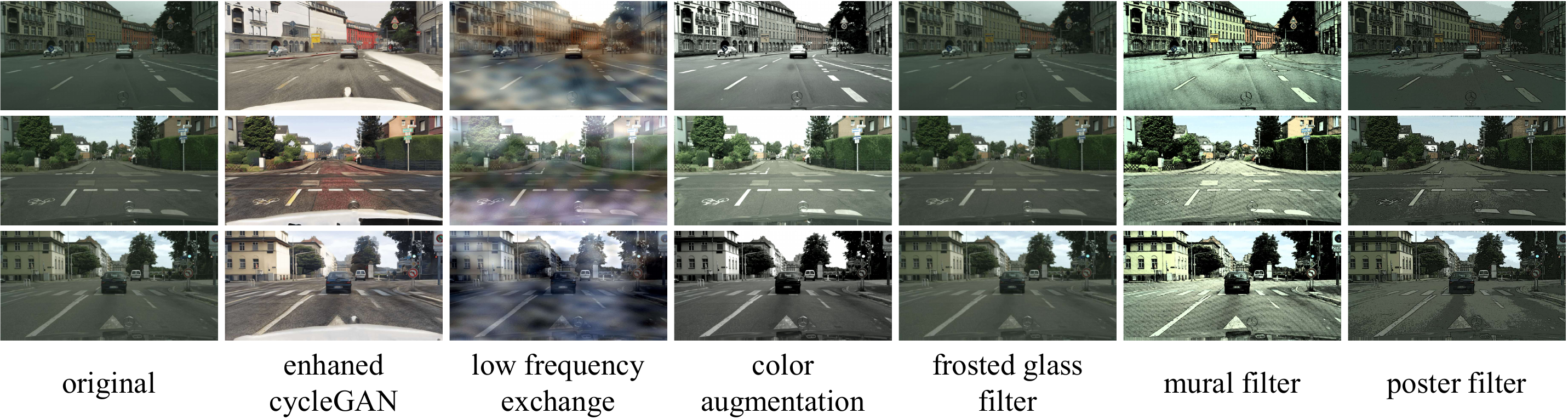} 
    \caption{Visualization of different image augmentation methods for adjusting for domain shift. Best viewed in color.}
    \label{dataaugmentation}
    \vspace{-1.0em}
\end{figure*}

Similarly, we envision that representation shift is also suitable for measuring domain shift in semantic segmentation tasks: for segmentation models trained on source domain data, the mIoU metric tested on the target domain is also highly correlated with representation shift. We carry out relevant experiments to prove this. Below follows a specific description of the representation shift used in our experiments:
\begin{figure}[t]
    \centering
    \includegraphics[width=0.99\columnwidth]{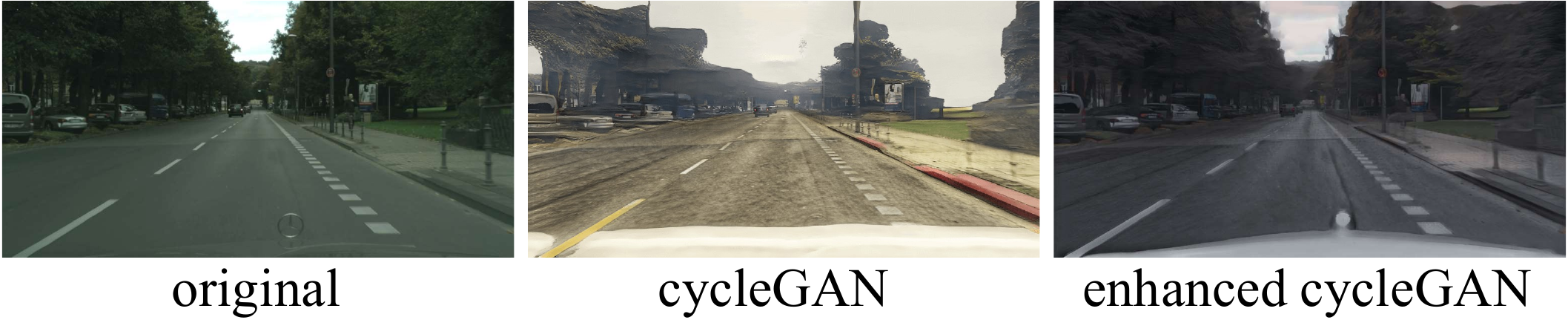} 
    \caption{Compared with the original cycleGAN, enhanced cycleGAN can preserve more content information.}
    \label{enhancedcyclegan}
    \vspace{-1.2em}
\end{figure}
consider a semantic segmentation model with a feature extraction module and a classification module. Let $F(x)=\left \{ f_1(x),...,f_c(x) \right \}$ denote the extracted feature map of the input image $x$, where $f_c(x)\in \left \{ \mathbb{R}^{h\times w } \right \}$ represents the $c$ th channel of the feature map. The average value of $f_c(x)$ is denoted as $a_c(x)$:

\begin{equation}
a_c(x)=\frac{1}{h} \frac{1}{w} \sum_{i,j}^{h,w} f_c(x)_{i,j}.
\end{equation}
Let $p_{a_{c} } ^{\mathcal{S} }$ denotes the continuous distribution of $a_c(x)$ values across the source domain dataset $X ^{\mathcal{S} } =\left \{ x_{1}^{\mathcal{S}},...,x_{n_s}^{\mathcal{S}} \right \}$, where $n_s$ is the number of images in source domain. Similarly, the distribution $p_{a_{c} } ^{\mathcal{T} } $ of the target domain dataset $X ^{\mathcal{T} } =\left \{ x_{1}^{\mathcal{T}},...,x_{n_t}^{\mathcal{T}} \right \} $ can be obtained.

Then the representation shift $R$ for semantic segmentation is defined as follows:
\begin{equation}
\label{R}
    R\left ( X^{\mathcal{S}},X^{\mathcal{T}} \right )=\frac{1}{c} \sum_{i=1}^{c} W(p_{a_{i} } ^{\mathcal{S}},p_{a_{i} } ^{\mathcal{T} }),
\end{equation}
where $W$ denotes the Wasserstein distance between the two distributions.

\subsection{Feasibility Analysis}

\begin{figure*}[t]
\centering
\subfigure[WSSS-UDA: Weakly Supervised Semantic Segmentation + Unsupervised Domain Adaptation]{
\label{scheme1}
\includegraphics[width=0.90\textwidth]{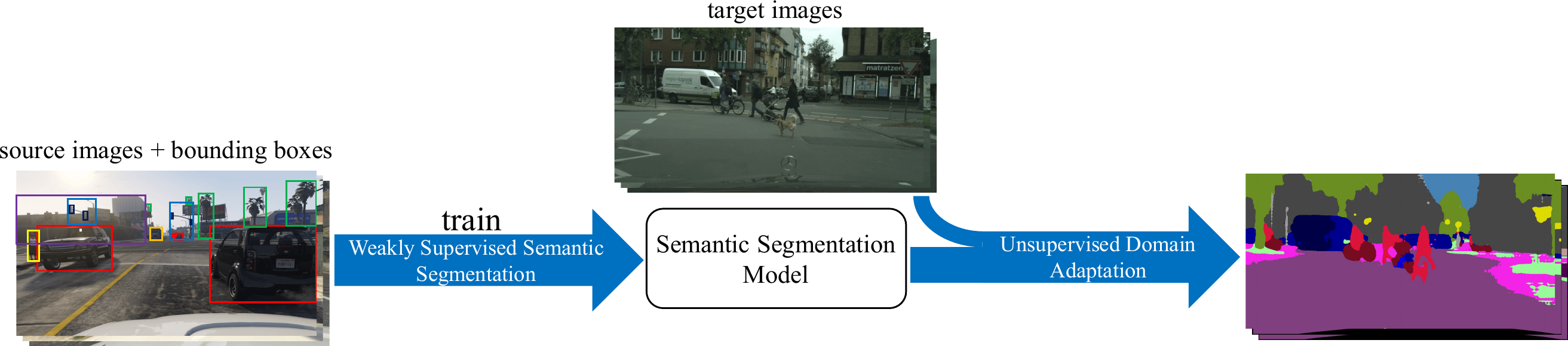}}
\subfigure[TDOD-WSSS: Target Domain Object Detection + Weakly Supervised Semantic Segmentation]{
\label{scheme2}
\includegraphics[width=0.95\textwidth]{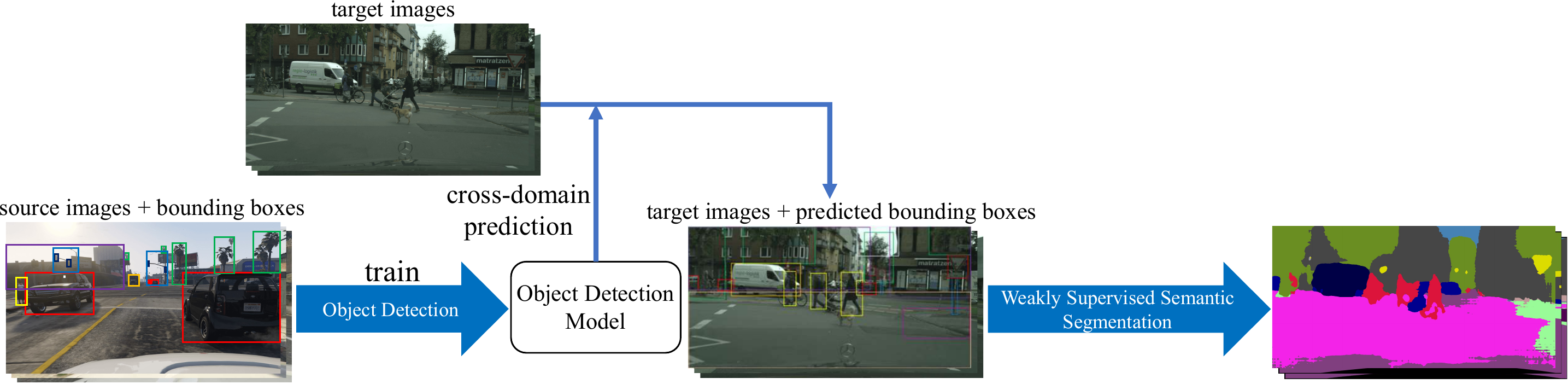}}
\caption{Frameworks for WUDA task. (a) First, perform box-supervised semantic segmentation in the source domain. With the segmentation model pre-trained on the source domain, UDA methods can be performed in the target domain to enhance the generalization ability of the model. (b)  Train an object detection model in the source domain. With a pre-trained object detection model in the source domain, we can predict bounding boxes on the target domain and then implement weakly supervised segmentation.}
\label{schemes}
\vspace{-1.0em}
\end{figure*}

To verify the rationality of applying representation shift in image segmentation tasks, we use four common semantic segmentation datasets for our research: Cityscapes \cite{cordts2016cityscapes}, BDD100k \cite{yu2020bdd100k}, GTAV \cite{richter2016playing} and SYNTHIA \cite{ros2016synthia}. Each measurement of representation shift uses one of the datasets as the source domain and the other three datasets as target domains (e.g. GTAV as the source domain, cityscapes, SYNTHIA and BDD100k as the target domain). 

Our data collection consists of the following 3 steps: 1) Train a semantic segmentation model using source domain data. 2) Use the trained model to extract the features of the source and target domains, then calculate the representation shift defined in Eq.~\ref{R}. 3) Use the trained model to make predictions in the target domain and calculate mIoU. During the training process in step 1, we use a Deeplabv3 for semantic segmentation, set the batch size to 2, and perform 10,000 iterations without using any data enhancement methods such as resize, flip, and clipping to ensure that the model fits the original images in the source domain. The collected representation shift-mIoU pairs are used for linear regression analysis.

As can be seen from Figures ~\ref{Fig.sub.1} and ~\ref{Fig.sub.2}, the value of representation shift and mIoU have a high degree of negative correlation. Pearson correlations of representation shift and mIoU reach -0.874 and -0.741 for 16 and 19 categories semantic segmentation, respectively. This conclusion is similar to that of the pathological image classification task in the article \cite{stacke2020measuring}, which proves that for the semantic segmentation task, using the representation shift to measure the domain shift is also applicable.

\subsection{Data Construction}

For the popularly used dataset pair GTAV-Cityscapes in UDA methods, we perform different data augmentation processes on the target domain (Cityscapes) to simulate different domain shifts. This helps in section \textbf{Experiments} to analyze the impact of different domain shifts on our proposed frameworks. The visualization of image augmentations is shown in Figure~\ref{dataaugmentation}. For possible future needs to construct dataset pairs with specific domain shifts, we propose a data construction process (Algorithm~\ref{alg:algorithm}). Assume that data augmentation operations are sufficient.

\begin{algorithm}[tb]
\caption{Dataset Construction Algorithm}
\label{alg:algorithm}
\textbf{Input}: expected domain shift interval $(a-\delta ,a+\delta )$, source domain dataset $X ^{\mathcal{S} }$, target domain dataset $X ^{\mathcal{T} }$, a list of image augmentation operations $\left \{ O_1,...O_n  \right \} $, domain shift calculation formula $R$.\\
\textbf{Output}: expected dataset $\hat{X} ^{\mathcal{T} }$.
\begin{algorithmic}[1] 
\STATE Let $i=1$.
\WHILE{$i<=n$}
\STATE $\hat{X} ^{\mathcal{T}}=O_i(X ^{\mathcal{T} })$.
\IF {$R(\hat{X} ^{\mathcal{T}},X ^{\mathcal{S}})$ \textbf{in} $(a-\delta ,a+\delta )$}
\STATE \textbf{break}
\ELSE
\STATE $i=i+1$.
\ENDIF
\ENDWHILE
\STATE \textbf{return} $\hat{X} ^{\mathcal{T}}$
\end{algorithmic}
\end{algorithm}

~\\
\noindent
\textbf{Enhanced cycleGAN}
CycleGAN \cite{zhu2017unpaired} is a common method for unsupervised image-to-image translation, which can realize the conversion of image style between two domains. In order to improve the generalization ability of the model, cycleGAN is used in many studies to reduce the domain shift. Therefore, we implement an enhanced cycleGAN to transfer Cityscapes images into GTAV style. Original cycleGAN only performs adversarial training at the image level, which may result in missing content in the style-transferred Cityscapes images. We add a discriminator on each transfer branch (source to target/target to source) to conduct class-level adversarial training: for generated images, the additional discriminator identifies each part (e.g. vegetation, building, etc.) of the image is real or not. As shown in Figure~\ref{enhancedcyclegan}, our enhanced version of cycleGAN can further preserve the content information of the image while changing the style of the image.

~\\
\noindent
\textbf{Low frequency exchange}
FDA \cite{yang2020fda} introduces the method of low frequency exchange in unsupervised domain adaptation for semantic segmentation. Perform fast Fourier transform (FFT) on the source domain images and target domain images separately and transplants the low frequency parts of the source domain to the target domain, so that the target domain has the low-frequency features of the source domain. Theoretically, low frequency exchange can bring the distribution of the target domain closer to the source domain.

~\\
\noindent
\textbf{Other image augmentations}
We also made other augmentations to target domain images: 1) Color augmentation. Randomly adjust the brightness, contrast and color of the image, as shown in Figure~\ref{dataaugmentation}, column 4; 2) Image filters. Use image processing software to filter images with the following effects: frosted glass,  mural and poster, as shown in Figure~\ref{dataaugmentation}, columns 5-7. These image augmentation methods change the domain shift between the source and target domains to varying degrees.

~\\
The constructed datasets cover a wide range of domain shifts as shown in Figure~\ref{Fig.sub.3}. The relationship between mIoU and representation shift further justifies the use of representation shift to measuring domain shift in semantic segmentation tasks.

\begin{table*}[htbp]
\centering
\footnotesize
\setlength\tabcolsep{2.4pt}{
\begin{tabular}{c|*{19}{c}|c}
\toprule 
 & \multicolumn{1}{c}{\begin{sideways}Road\end{sideways}} & \multicolumn{1}{c}{\begin{sideways}SW\end{sideways}} & \multicolumn{1}{c}{\begin{sideways}Build\end{sideways}} & \multicolumn{1}{c}{\begin{sideways}Wall\end{sideways}} & \multicolumn{1}{c}{\begin{sideways}Fence\end{sideways}} & \multicolumn{1}{c}{\begin{sideways}Pole\end{sideways}} & \multicolumn{1}{c}{\begin{sideways}TL\end{sideways}} & \multicolumn{1}{c}{\begin{sideways}TS\end{sideways}} & \multicolumn{1}{c}{\begin{sideways}Veg.\end{sideways}} & \multicolumn{1}{c}{\begin{sideways}Terrain\end{sideways}} & \multicolumn{1}{c}{\begin{sideways}Sky\end{sideways}} & \multicolumn{1}{c}{\begin{sideways}Person\end{sideways}} & \multicolumn{1}{c}{\begin{sideways}Rider\end{sideways}} & \multicolumn{1}{c}{\begin{sideways}Car\end{sideways}} & \multicolumn{1}{c}{\begin{sideways}Truck\end{sideways}} & \multicolumn{1}{c}{\begin{sideways}Bus\end{sideways}} & \multicolumn{1}{c}{\begin{sideways}Train\end{sideways}} & \multicolumn{1}{c}{\begin{sideways}Motor\end{sideways}} & \multicolumn{1}{c}{\begin{sideways}Bike\end{sideways}} & \multicolumn{1}{|l}{mIoU}
 \\
\midrule
 CBST (fine label) & 91.8 & 53.5& 80.5& 32.7& 21.0& 34.0& 28.9& 20.4& 83.9& 34.2& 80.9& 53.1& 24.0& 82.7& 30.3& 35.9& 16.0& 25.9& 42.8& 45.9 \\
 E-IAST (fine label)&94.6 & 65.9 & 87.6 & 41.4& 25.8& 36.4& 52.1& 54.8& 82.7& 23.0& 89.1& 68.0& 23.6& 88.2& 42.4& 52.5& 26.6& 50.8& 62.0& 56.2\\
\midrule
WSSS-UDA (CBST)&27.3 & 20.3 & 48.7 & 11.9& 9.2& 10.1& 20.1& 11.3& 65.1& 23.6& 46.0& 50.1& 15.0& 73.1& 16.5& 6.2& 0.0& 23.2& 26.1& 26.5\\
WSSS-UDA (E-IAST)& 82.5 & 39.1 & 82.4 & 34.4 & 34.2 & 21.0 & 44.6 & 43.4 & 76.7 & 20.4 & 63.4 & 63.8 & 6.5 & 83.2 & 31.7 & 47.5 & 0.0 & 46.5 & 60.0 & 46.4\\
TDOD-WSSS (Yolov5)&61.9 & 21.1 & 76.3 & 28.3 & 26.8 & 6.7 & 38.2 & 28.4 & 74.3 & 14.4 & 59.5 & 57.0 & 11.4 & 79.3 & 20.7 & 23.0 & 0.0 & 0.0 & 0.3 & 33.0\\
\bottomrule  
\end{tabular}
}
\caption {mIoU (\%) results of semantic segmentation on Cityscapes with weak GTAV labels (synthesis$\to$real). The remarks behind Framework WSSS-UDA represent which UDA method is used. The remark behind Framework TDOD-WSSS represents which object detection method is used. `Fine label' means that precise semantic labels are used in the source domain.}
\label{tab:Comparison results from GTA5 to Cityscapes}%
\vspace{-1.0em}
\end{table*}

\begin{table*}[htbp]
\centering
\footnotesize
\setlength\tabcolsep{2.4pt}{
\begin{tabular}{c|*{19}{c}|c}
\toprule 
 & \multicolumn{1}{c}{\begin{sideways}Road\end{sideways}} & \multicolumn{1}{c}{\begin{sideways}SW\end{sideways}} & \multicolumn{1}{c}{\begin{sideways}Build\end{sideways}} & \multicolumn{1}{c}{\begin{sideways}Wall\end{sideways}} & \multicolumn{1}{c}{\begin{sideways}Fence\end{sideways}} & \multicolumn{1}{c}{\begin{sideways}Pole\end{sideways}} & \multicolumn{1}{c}{\begin{sideways}TL\end{sideways}} & \multicolumn{1}{c}{\begin{sideways}TS\end{sideways}} & \multicolumn{1}{c}{\begin{sideways}Veg.\end{sideways}} & \multicolumn{1}{c}{\begin{sideways}Terrain\end{sideways}} & \multicolumn{1}{c}{\begin{sideways}Sky\end{sideways}} & \multicolumn{1}{c}{\begin{sideways}Person\end{sideways}} & \multicolumn{1}{c}{\begin{sideways}Rider\end{sideways}} & \multicolumn{1}{c}{\begin{sideways}Car\end{sideways}} & \multicolumn{1}{c}{\begin{sideways}Truck\end{sideways}} & \multicolumn{1}{c}{\begin{sideways}Bus\end{sideways}} & \multicolumn{1}{c}{\begin{sideways}Train\end{sideways}} & \multicolumn{1}{c}{\begin{sideways}Motor\end{sideways}} & \multicolumn{1}{c}{\begin{sideways}Bike\end{sideways}} & \multicolumn{1}{|l}{mIoU}
 \\
\midrule
 E-IAST (fine label)&95.1 & 67.2 & 86.7 & 34.0& 28.4& 39.9& 46.9& 52.0& 86.6& 45.0& 85.1& 66.4& 38.6& 89.4& 45.2& 52.0& 0.0& 49.5& 61.7& 56.3\\
\midrule
WSSS-UDA (E-IAST)& 78.5 & 35.0 & 81.8 & 31.4 & 21.0 & 16.0 & 40.7 & 51.8 & 82.3 & 27.6 & 76.1 & 60.7 & 10.8 & 84.7 & 35.8 & 41.1 & 0.0 & 44.4 & 58.7 & 46.9\\
TDOD-WSSS (Yolov5)&71.1 & 30.5 & 77.7 & 26.6 & 31.8 & 8.0 & 18.0 & 39.8 & 78.6 & 25.0 & 59.9 & 57.0 & 30.2 & 80.2 & 48.2 & 45.1 & 0.0 & 34.2 & 54.9 & 43.0\\
\bottomrule  
\end{tabular}
}
\caption {mIoU (\%) results of semantic segmentation on Cityscapes with weak BDD100k labels (real$\to$real).}
\label{tab:Comparison results from BDD to Cityscapes}%
\vspace{-1.0em}
\end{table*}

\section{Frameworks}

Considering that WUDA contains both weakly supervised and cross-domain problems, we first propose the framework WSSS-UDA: In order to take advantage of the fine bounding box labels in the source domain, first perform box-supervised segmentation in the source domain, then the problem is transformed into a UDA task. The schematic diagram of the framework WSSS-UDA is shown in Figure~\ref{scheme1}.

However, the cross-domain process is not necessarily done on the semantic segmentation task. According to the study \cite{redmon2016you}, Yolo has a strong generalization ability: when trained on natural images and tested on the artwork, Yolo outperforms the two-stage detection methods (e.g. R-CNN \cite{girshick14CVPR}) by a wide margin. Therefore, we implement the cross-domain process on the object detection task and propose the framework TDOD-WSSS (Figure~\ref{scheme2}): First, use the bounding box labels of the source domain to train the object detection model, then predict bounding boxes on the target domain. Finally, implement box-supervised segmentation in the target domain.

We benchmark typical methods under our proposed frameworks. For weakly supervised semantic segmentation, we first use GrabCut to obtain pseudo-labels, then perform three epochs of self-training to gradually improve the accuracy of the segmentation results. For unsupervised domain adaptation, we benchmark CBST \cite{zou2018unsupervised} and enhanced IAST \cite{mei2020instance}, where CBST is an ST-based method and IAST combines ST and AT. Our enhanced IAST optimizes the self-training process of the original IAST. For the object detection method, we adopt the widely used Yolov5 \cite{glenn_jocher_2020_3983579}.

\section{Experiments} \label{experiments}
This section will introduce the datasets and metrics, the implementation details of the experiments, and the presentation and analysis of the results.

\begin{figure*}[t]
    \centering
    \includegraphics[width=0.99\textwidth]{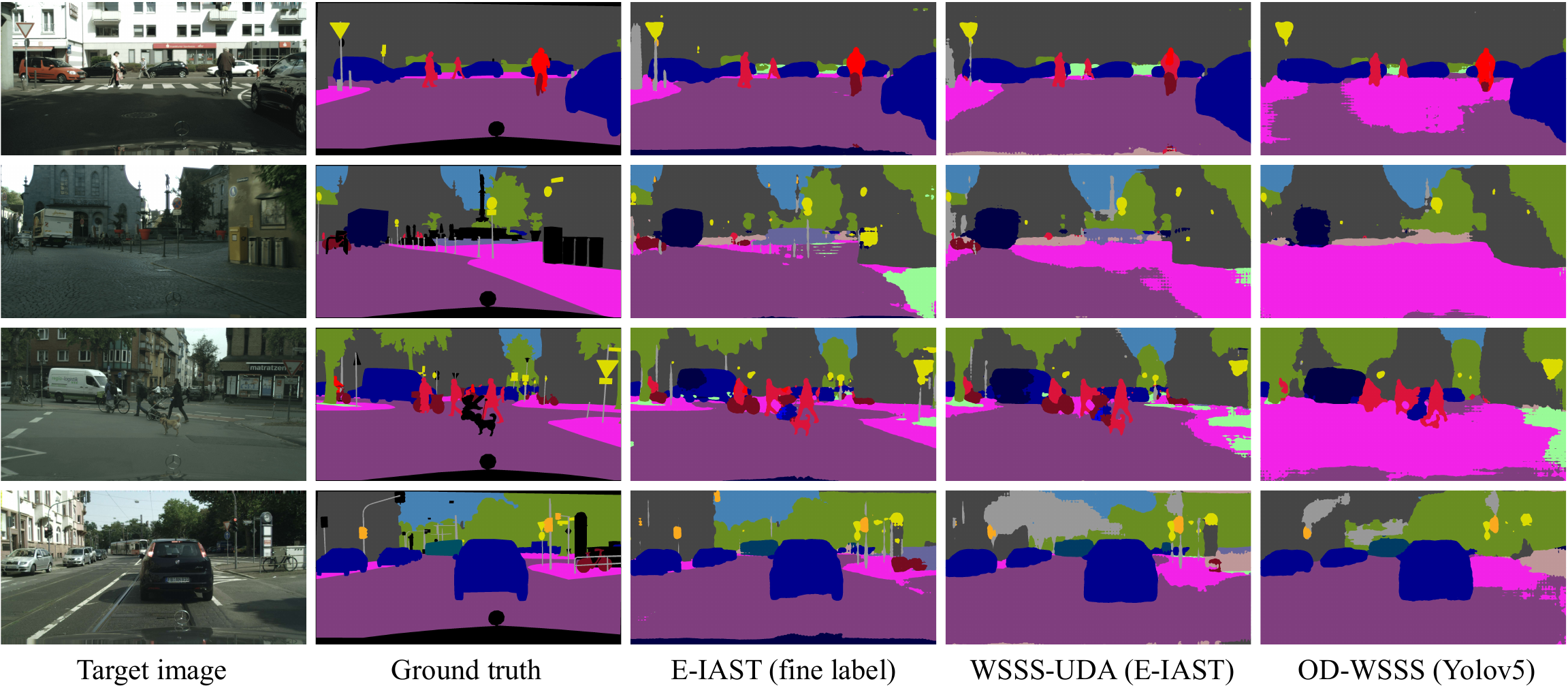} 
    \caption{Visualization of Semantic Segmentation on Cityscapes with different frameworks. Framework WSSS-UDA can achieve results close to UDA methods with fine source labels. However, the results of framework TDOD-WSSS are rough in detail and have many wrong segmentations. Best viewed in color.}
    \label{resultsshow}
    \vspace{-1.0em}
\end{figure*}

\subsection{Datasets and Evaluation}
GTAV \cite{richter2016playing} and Cityscapes \cite{cordts2016cityscapes} are the most widely used autonomous driving datasets in UDA methods. For WUDA, we also adopt the GTAV-Cityscapes dataset pair in our experiments. In addition, in order to simulate the real-to-real scene, we carried out further experiments on the dataset pair BDD100k-Cityscapes.

GTAV dataset as the source domain does not have the annotation of the bounding boxes. We use the python package scikit-image to perform class-wise connected domain detection on the semantic labels of GTAV, and then obtain the box of each connected domain.

We use the GrabCut method in the OpenCV package to extract pseudo-labels from bounding boxes. When an occlusion occurs between objects, we assume that the smaller target is in front, as in the method \cite{khoreva2017simple}.

For metrics, we use mean intersection over union (mIoU) for evaluation in all experiments.

\subsection{Implementation}
All our experiments are implemented with Pytorch on a single NVIDIA Tesla V100 with 32 GB memory. In framework WSSS-UDA, to ensure model compatibility, the model used for weakly supervised segmentation is consistent with the corresponding UDA method: CBST uses Deeplabv2 \cite{chen2017deeplab} with ResNet-38 \cite{he2016deep} as the backbone and our enhanced IAST uses Deeplabv2 with ResNet-101 as the backbone. All other settings remain the same as the default setting for CBST and IAST. In framework TDOD-WSSS, the weakly supervised segmentation step uses Deeplabv2 with ResNet-101 as the backbone. The object detection step uses a randomly initialized Yolov5l.

\subsection{Main Results}

The results of the two frameworks for WUDA are shown in Table \ref{tab:Comparison results from GTA5 to Cityscapes} and \ref{tab:Comparison results from BDD to Cityscapes}.
On the synthesis-to-real dataset, the results show that framework WSSS-UDA with E-IAST has a higher potential, its highest mIoU can reach 46.4\%, and this value reaches 82.6\% of E-IAST with fine source domain labels. However, for WSSS-UDA with CBST, the weak source domain labels bring a relatively large attenuation to the results (45.9\% $\to$ 26.5\% ). Framework TDOD-WSSS only achieves a mIoU of 33.0\%. This is because the Yolov5 trained on the source domain has misdetections on the target domain.

On the real-to-real dataset, the mIoU result of framework WSSS-UDA reaches 46.9\%, which is similar to that on the synthesis-to-real dataset. For framework TDOD-WSSS, the result is 43\%, which is a 10 percentage point increase compared to the result on the GTAV-Cityscapes dataset. As the dataset changes, the results of the two frameworks have very different trends. According to Figure\ref{Fig.sub.1}, the domain shifts of GTAV-Cityscapes and BDD100k-Cityscapes are significantly different. We can reasonably hypothesize that the two frameworks have different sensitivities to changes  of domain shifts. Therefore, we conduct extended experiments to verify our hypothesis in the next subsection.

Figure \ref{resultsshow} shows the qualitative results of semantic segmentation on Cityscapes. With the support of accurate labels in the source domain, the enhanced IAST can achieve very accurate segmentation, and the segmentation error area is very small. When using bounding boxes as the source domain labels, framework WSSS-UDA also achieves generally satisfactory results, however, it is not as detailed as the fully supervised UDA. When it comes to framework TDOD-WSSS, the segmentation results become coarser and have large areas of mis-segmented regions. This is because the misdetection of Yolov5 brings more noise to the subsequent weakly supervised segmentation.

We also employed the two-stage method faster-RCNN for object detection, however, the misdetection of this model in the target domain is catastrophic. Therefore, we did not implement further experiments with two-stage object detection methods.

\subsection{Analysis under Multiple Domain Shifts}

\begin{table}[htb]
\begin{center}{
\resizebox{0.99\linewidth}{!}{
\begin{tabular}{cccc}
\toprule
                    target domain style  & \makecell{WSSS-UDA\\(E-IAST)}  & \makecell{WSSS-UDA\\(CBST)}   &\makecell{TDOD-WSSS\\(Yolov5)}\\ \midrule
original                  & 46.4    & 26.5   & 33.0  \\
enhanced cycleGAN         & 42.0    & 26.7   & 28.7  \\ 
low frequency exchange    & 41.6    & 24.2   & 33.4  \\ 
color augmentation        & 43.0    & 24.5   & 25.7  \\
mural filter              & 37.6    & 23.5   & 6.0   \\ 
frosted glass filter      & 35.2    & 13.4   & 3.7   \\
poster filter             & 39.6    & 20.4   & 1.9   \\ 
\bottomrule
\end{tabular}}}
\end{center}
\caption{mIoU (\%) results of semantic segmentation on datasets with multiple domain shifts (the source domain dataset is GTAV).} 
\label{multi-style_mIoU}
\vspace{-1.0em}
\end{table}

\begin{figure}[t]
    \centering
    \includegraphics[width=0.99\columnwidth]{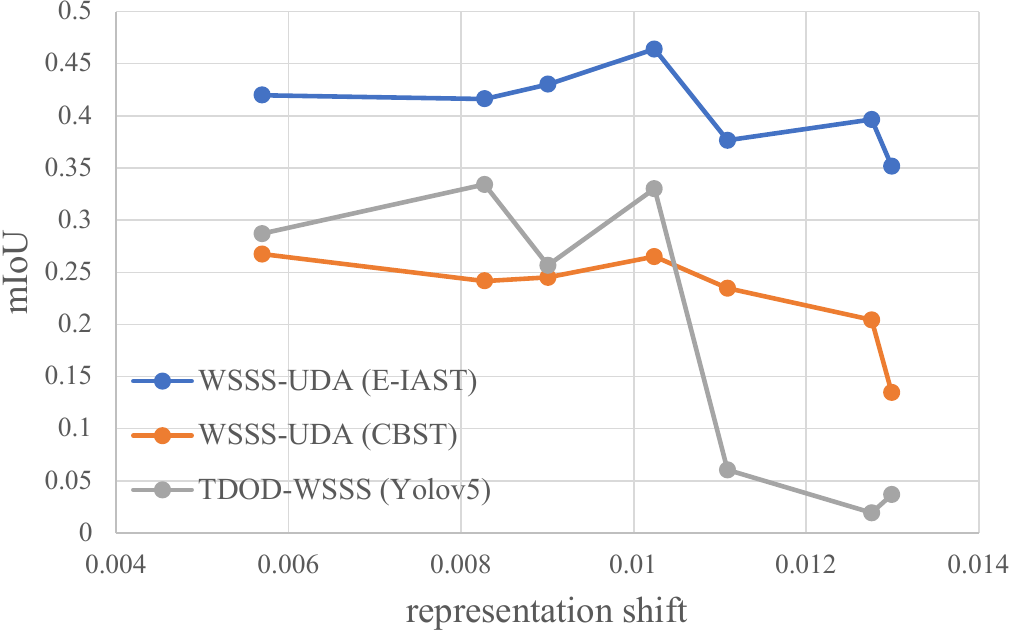} 
    \caption{The variation of mIoU results of different frameworks with domain shift.}
    \label{domainshiftanalysis}
    \vspace{-1.0em}
\end{figure}
~\\
Figure~\ref{domainshiftanalysis} shows the variation of mIoU results of different frameworks with domain shift. When the domain shift is moderate or small, the mIoU of WSSS-UDA is more stable than that of TDOD-WSSS, and WSSS-UDA with the E-IAST method can stabilize at a high level. When it comes to large domain shift, the mIoU of both frameworks drops: WSSS-UDA drops slowly and TDOD-WSSS drops significantly. The mIoU values of different frameworks under different domain shifts are shown in Table~\ref{multi-style_mIoU}.

In general, the domain shift has a larger impact on TDOD-WSSS, while WSSS-UDA is less sensitive to changes in domain shift. WSSS-UDA has more potential in the applications of complex target domains.
\section{Conclusion}

This paper defines a novel unsupervised domain adaptation task that requires only source domain bounding boxes for supervision. Meanwhile, we also propose two intuitive frameworks for this task. The results show that by using a suitable UDA method in the framework WSSS-UDA, mIoU on the target domain can reach 83\% of UDA methods with fine source labels. In addition, we apply representation shift to semantic segmentation of urban landscapes for the first time and analyze the impact of different domain shifts on the two proposed frameworks. Experiments prove that framework WSSS-UDA is more tolerant of domain shift.

However, there is still a long way to go for current methods to achieve precise segmentation on WUDA tasks. We hope more excellent solutions will be proposed to tackle WUDA in the future.

\bibliography{aaai22}
\end{document}